# Coalition-based Planning of Military Operations: Adversarial Reasoning Algorithms in an Integrated Decision Aid

Larry Ground
Alexander Kott
Ray Budd

BBN Technologies
5 PPG Place, Ste 310
Pittsburgh, PA 15222
{lground, akott, rbudd}@bbn.com

Abstract. Use of knowledge-based planning tools can help alleviate the challenges of planning a military operation in a coalition environment. We explore these challenges and potential contributions of knowledge-based tools using as an example the CADET system, a knowledge-based tool capable of producing automatically (or with human guidance) battle plans with realistic degree of detail and complexity. In ongoing experiments, it compared favorably with human planners. Interleaved planning, scheduling, routing, attrition and consumption processes comprise the computational approach of this tool. From the coalition operations perspective, such tools offer an important aid in rapid synchronization of assets and actions of heterogeneous assets belonging to multiple organizations, potentially with distinct doctrine and rules of engagement. In this paper, we discuss the functionality of the tool, provide a brief overview of the technical approach and experimental results, and outline the potential value of such tools for coalition operations.

## 1. Overview

Influential voices in the US military community (Wass de Czege and Biever, 2001) argue for significant computerization of the military planning process and for "...fast new planning processes that establish a new division of labor between man and machine. Staffs will rely heavily upon software to complete the straightforward calculations. Decision aids will quickly offer suggestions and test alternative courses of actions." Although the reasons for introducing such a computerization in the military planning processes are compelling enough even in the context of a single-nation military, many of the same reasons become even more pronounced in a coalition environment:

- The process of planning a military operation remains relatively cumbersome, inflexible and slow even when conducted by a planning staff that trained together extensively in order to achieve common understanding of the collaborative procedures, approaches and ontology. In a coalition context, the planning staff rarely has the benefits of extensive joint training, and comes into the process with significantly different sets of procedures, terminology, and doctrines (Riscassi, 1993).

- The planning process frequently involves significant disagreements on estimation of outcomes, attrition, consumption of supplies, and enemy reactions. Much of these disagreements arise from differences in mental models and underlying assumptions of the process participants. Such differences are further exacerbated in planning performed by a coalition staff (Elron et. al., 1999).

- There is a fundamental complexity of synchronization and effective utilization of multiple heterogeneous assets performing numerous, inter-dependent, heterogeneous tasks. This complexity, heterogeneity and the need for careful coordination and synchronization inevitably grow in a coalition environment, particularly for the ground component.

We argue that using an effective decision aid can, in part, alleviate these challenges. As an example, consider CADET, a tool for producing automatically (or with human guidance) Army battle plans with realistic degree of detail and complexity. In ongoing experiments, it compared favorably with human planners.

In brief, the human planner defines the key goals for a tactical course of action (COA), and CADET expands them into a detailed plan/schedule of the operation. CADET expands friendly tasks, determines the necessary supporting relations, allocates / schedules tasks to friendly assets, takes into account dependencies between tasks and availability of assets, predicts enemy actions and reactions, devises friendly counter-actions, estimates paths of movements, timing requirements, attrition and risk. CADET is a generic engine, not specific to any type of assets or tasks. Although currently it is fitted with a US Army-specific task model, it can be readily augmented with models for other forces and nations, a clear requirement for coalition warfare.

Recently, there were several efforts to utilize the planning capability introduced by CADET. For example, US Army Battle Command Battle Lab-Leavenworth (BCBL-L) chose CADET as the centerpiece for its Integrated COA Critiquing and Evaluation System (ICCES) program to provide task expansion for maneuver COAs created with sketching tools and plan developers.

DARPA applied CADET in its Command Post of the Future (CPoF) program as a tool to provide a maneuver course of action. Under the umbrella of the CPoF program, CADET was integrated with the FOX GA system (Hayes and Schlabach, 1998) to provide a more detailed planner, coupled with COA generation capability. Battle Command Battle Lab-Huachuca (BCBL-H) integrated CADET with All Source Analysis System-Light (ASAS-L) to provide a planner for intelligence assets and to wargame enemy COAs against friendly COAs.

The development of Course of Action Display and Evaluation Tool (CADET) began in 1996, at the Carnegie Group, Inc. under the funding available under the Small Business Innovative Research (SBIR) program. With numerous other efforts addressing various aspects of the military decision-making process (MDMP), we sought to concentrate our efforts on the COA analysis phase of the MDMP.

In a setting such as a US Army divisional planning cell, the detailed analysis of a tactical course of action involves a staff of 3-4 persons with in-depth knowledge of both friendly and enemy tactics. Working as a team, they ascertain the feasibility of the COA, to assess its likelihood of success against a particular enemy COA, and to identify the points of the COA requiring synchronized action for participants. The resulting analysis is usually recorded in a matrix format, with time periods for the columns and functional alignment, such as the Battlefield Operating Systems (BOS), for the rows (Field Manual 101-5). Comparable, although not necessarily identical elements exist in decision-making processes of other nations' military establishments, and will be undoubtedly found, formally or informally, in any coalition decision-making.

## 2. Challenges and Capabilities

A planning tool for coalition warfare must provide numerous capabilities to address a number of key challenges. Such capabilities fall into several broad categories:

- Modeling of assets and tasks
- Adversarial environment
- Coordinating team efforts
- Autonomous action

In this section, we explore some examples of such capabilities and their possible relations to coalition operations, from a functional, domain-oriented perspective.

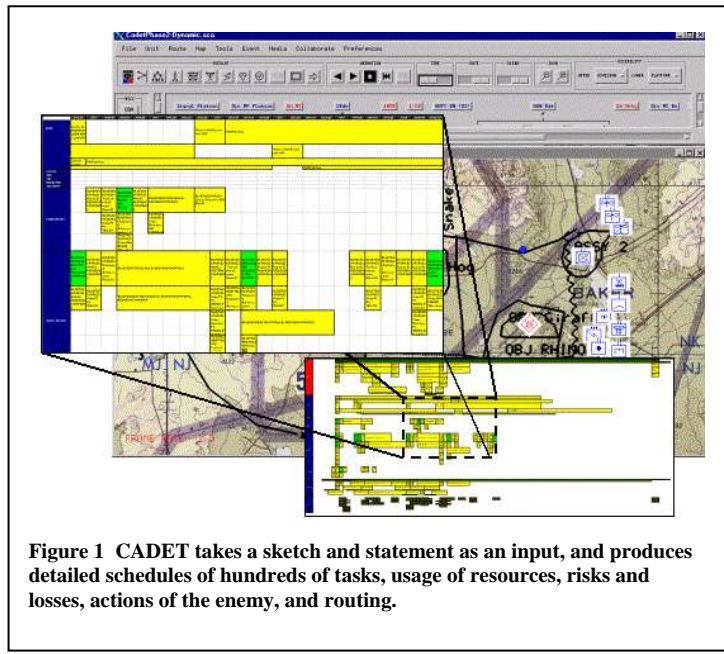

**Figure 1** CADET takes a sketch and statement as an input, and produces detailed schedules of hundreds of tasks, usage of resources, risks and losses, actions of the enemy, and routing.

### *Modeling of assets and tasks*

Coalitions bring together military assets with different capabilities and employment doctrines. All too often, a coalition includes members whose assets, capabilities and tactics are not particularly familiar to other members. Thus, any decision aid for coalition planning must allow flexible, inexpensive, and rapid modeling of assets and associated tasks.

Let us consider the evolution of modeling the air assets in CADET as an example. Initially, we started with a very simple modeling that calculated deployment/re-deployment times and time-on-station, but with flat rates applied to resource consumption and timing considerations.

As the modeling evolved, we captured the variations caused by a variety of different aspects of the employment cycle. For example, working with the Battle Command Battle Lab-Huachuca, we performed a detailed breakdown of the sub-tasks involved in readying, launching and positioning a UAV. The possibility of concurrent tasks was factored in where the UAV could be routed to collect intelligence along the ingress/egress route.

The impact of the UAV use on the ground maneuver plan was greater than originally expected. Subject matter experts (SME) had predicted the ground commander would use the UAVs primarily to verify *known* or *suspected* information. Further analysis revealed a prejudice toward UAVs by the older generation based on experience with weather-constrained Army aviation and a tendency to focus on operations within their immediate control. Younger officers, however, employed UAVs as a primary source for intelligence, integrating them fully into the intelligence collection plan.

CADET added a new dimension to the modeling of UAV by showing the demands of continuous coverage. Users had generally planned individual missions or multiple missions. Few of them had considered the full implications of putting continuous coverage on a target. Army attack helicopters address this by using one of three modes: attack by platoon, attack by company or simultaneous attack.

A unit with a limited number of UAVs must factor in travel and recovery time for the cycling UAVs to determine if continuous coverage is feasible. Users were generally discounting the cost of the recovery time (for refueling and preventive maintenance) when calculating the amount of time the UAVs were effectively available for on-site observation.

As this example illustrates, an approach to modeling of assets must take into account at least the following considerations: (a) it must provide for rapid, inexpensive insertion of an initial, coarse but serviceable model; (b) allow for gradual increase in the model's fidelity, with incremental modifications even in a field environment, and (c) recognize and accommodate significant differences between organizations, as well as the ongoing evolution, in approaches to asset's employment.

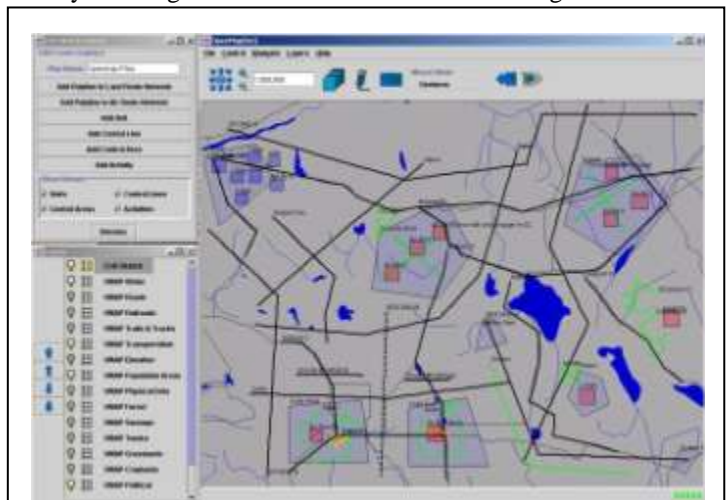

**Figure 2 One of the COA-editing tools that have been used as data-entry interfaces to CADET and an example of a sketch produced with the tool.**

*Adversarial environment*

Assumptions and expectations regarding the enemy are particularly challenging in a coalition, where the doctrine of staff officers from multiple nations can differ significantly and the political and strategic aims of the participating nations may be at odds (Riscassi 1993).

Manual wargaming typically depicts the enemy in a situation template, literally a standard tactical formation adapted to a specific piece of terrain in a given situation. Modeling the enemy over time, then, is a matter of taking the standard formations and moving them along the avenues of approach toward the friendly force.

In practice, there are several aspects in considering how the enemy affects friendly actions. In particular, every action taken by either combatant is likely to cause a reaction by the opponent and it might be possible to negate the reaction with the appropriate counteraction. Further, a quick, reliable Conflict Resolution Model (CRM) is needed to determine the effects of each engagement on the combatants.

**Action/reaction/counter-action**

Every action possible by either friendly or enemy units warrants examination for potential reactions. This is augmented with further analysis to determine if there exists a counter-action that can be used to minimize the impact of the reaction or negate its effects completely.

For example, whenever artillery is fired, the opposing force will attempt to locate the firing piece and fire counter-battery fire. The firing unit must either be prepared to relocate or expect to receive incoming fire. The general effect is to reduce harassing and interdicting fires whenever a credible counter-battery threat is present. The potential counter-action is for the firing unit that fired first to conduct counter-battery operations of its own. In fact, US forces have sometimes fired in hopes of drawing the enemy into counter-battery fire for the explicit purpose of destroying the enemy artillery through counter-battery fire. In a coalition environment, a planning tool must allow for multiple and readily adjustable models of such action-reaction-counteraction, to reflect diverse perspectives and expectations of the coalition members.

**Conflict Resolution Modeling (CRM)**

Although the approach of Dupuy (1990) offers many advantages for application in a system like CADET, the modest demands on the required data being one of them, we found that it produced results that were not in concert with those expected by the users. Having involved expert panels of military officers, both active duty and retired, we modified the equations and coefficients provided in (Dupuy, 1990) to match the expertise and experience of current practitioners (Kott, Ground and Langston, 1999). In a coalition planning process, it may be desirable to be able to either select from a library of multiple models, or to modify rapidly an existing one in a manner that takes into account the perspectives and experiences of the coalition members (Elron et. al., 1999).

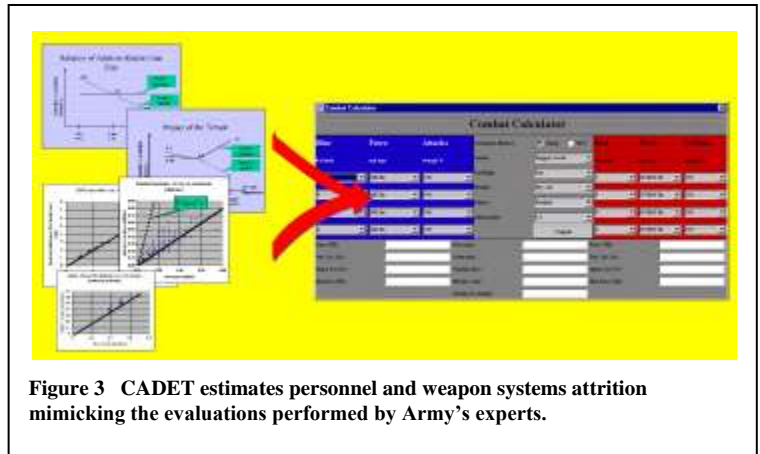

**Figure 3** CADET estimates personnel and weapon systems attrition mimicking the evaluations performed by Army's experts.

*Coordinating team efforts*

**Coordinating timing and movement**

Coalition warfare exacerbates the need for careful, thoughtful coordination of temporal and spatial aspects of all tasks within an operation. Field Manual 3-0, the US Army's keystone manual for operations, states "Detailed war-gaming, planning and rehearsals help develop a common understanding of the operation plan and control measures (Field Manual 3-0)." CADET's users can input temporal relationships for high-level activities for a plan. Subject matter expert and user feedback provided us with important information concerning the way a commander conceives the temporal relations between activities. For example, does an attack in an area start when the unit starts moving to the specified area, when the unit attacks the targeted unit, or when the unit enters the specified area?

In CADET, this problem is solved by identifying what we call anchor points for each activity. When the user says that two specified activities should start at the same time, he or she has a specific idea about which derived activities they want the units to be starting at the same time. Users were typically less concerned with the time at which a unit starts moving and more interested in each unit first makes contact with the enemy. For example, when performing a *Seize*, the start anchor point is the first movement in the area being seized. When performing a *Close-with-and-engage*, the start anchor point is the first attack on any target unit. In coalition operations, however, it is likely that officers from different doctrinal backgrounds will have different notions about such anchor points. This is yet another aspect of knowledge engineering in systems like CADET that requires a mechanism for rapid, in-field modifications.

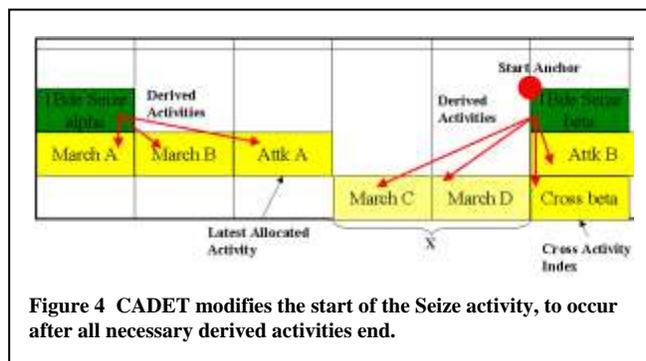

**Figure 4** CADET modifies the start of the Seize activity, to occur after all necessary derived activities end.

**Coordinating supporting relationships**

The common errors encountered in manual COA analysis include failure to fully utilize resources, committing resources to provide support when they are not within range, and over-committing resources.

Clearly, these errors would be even more likely to occur in a COA analysis process performed by a coalition staff. CADET's planning and scheduling algorithm ensures resources are allocated within constraints and are not over-committed. In those cases when the algorithm is unable to find a solution without an over-commitment of resources, CADET identifies the affected activity as questionable (e.g., Fig. 5), but continues the planning process. This allows the user to accept or correct the over-commitment of resources when a more complete solution is available for review and decision-making.

CADET tracks the utilization of resources to allow users to know where resources are not being fully exploited, a capability that can be used to look for places where resources could be applied elsewhere.

CADET looks at the effective range of supporting resources, such as logistics facilities, to determine if they are close enough to achieve the mission. For instance, CADET models the actual movement of support elements between the field trains and combat units. As the combat elements move forward in the offense, and the distances and the time required to perform re-supply increases as well. When it becomes too great to support the planned level of tactical operations, CADET cues the planner to reposition the field trains forward to a closer location. If the trains cannot be repositioned in a timely manner, CADET identifies the restrictions imposed on the combat unit by the reduced level of support. By taking care of such details, CADET can help the coalition staff avoid the typical mistakes of resource management in COA analysis.

**Operating in three dimensions**

In practice, human planners tend to focus exclusively on the close fight, without due consideration to the full depth of the battlespace. For example, leaders who lack experience with US Army attack helicopters tend to discount their value or leave them out of the equation completely.

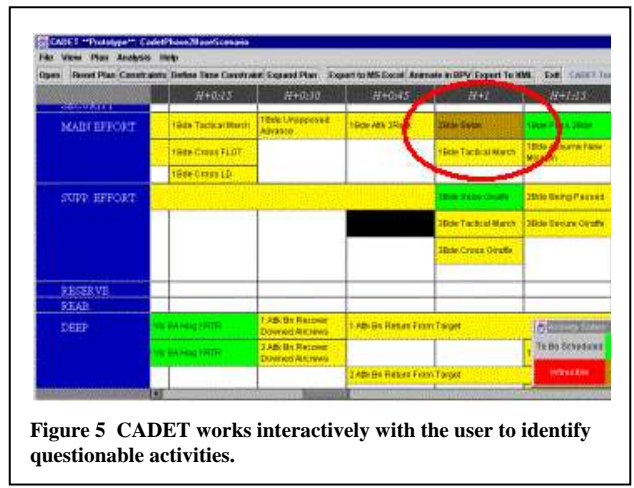

**Figure 5** CADET works interactively with the user to identify questionable activities.

A deep attack will normally cause serious attrition for the enemy but carries with it the risk of friendly losses. If Army attack helicopters are lost behind enemy lines, it necessitates a combat search and rescue (CSAR) mission. On the other hand, a deep attack could reduce the enemy strength to the point where the enemy is forced to call off the attack. Whenever assets are available, a deep attack should be considered. The coalition staff officers can take advantage of CADET's ability to analyze air attacks to build in COAs with air assets, where air and ground assets may belong to different coalition members.

*Autonomous action*

In the context of coalition warfare, even more so than in single-nation warfare, guidance from the commander should often come in the form of his intent or the desired results (Keithly and Ferris, 1999).

**Modeling tasks based on intent**

The bypass criterion in CADET provides the ability for units to disengage when the opposing force has been attrited to a certain level. However, it does not address the more general situation encountered where actions are initiated with a specific intent in mind. For instance, in *economy of force* operations, the supporting attack will generally not be able to destroy or even to defeat the enemy. Rather, the intent of the supporting attack is to ensure the success of the main effort, regardless of the extent to which the supporting effort is able to defeat the enemy.

Artillery fire commonly has an associated intent. Artillery will be used to suppress, to mask, to defeat, or to destroy. By extending the task set to include the intent, the applicability of the tasks to specific situations was greatly enhanced.

Modeling for deliberate attack is an excellent example of intent and its effect on resource consumption. In CADET, the task is modeled to allow the projection of attrition for attacks that are not attempting to completely remove the enemy (i.e. Attack to Attrit). The effect is a change to attack duration, and ultimately a modification to total defender and attacker attrition. For a planner, the need to hold the friendly strength at or above a certain threshold might be key to the analysis of a particular COA.

**Derived actions for subordinates based on higher level tasks**

A coalition operation consists of a large number of disparate, unique sub-tasks working to achieve a common goal. To properly model the task requires modeling a variable number of sub-tasks. The timing and interaction of the sub-tasks determines the success or failure of the task. Of particular interest is the assignment of tasks and routes to units that are not fully identified by the user.

A counter-attack is a good example. The commander will attempt to commit the counter-attack force at the time necessary to reverse the trend of the defense. The problem is that the exact speed and route of the attacking force can generally not be predicted in advance. The counter-attack force will be most effective if it is able to strike a flank.

CADET automatically calculates the route and timing for the counter-attack force's movement. In a deliberate planning mode, this allows time to perform route reconnaissance. In a real-time execution-replanning cycle, the ability to rapidly calculate routes and related timing would facilitate identification of the decision point for commitment.

Movement to contact, another good example, represents a significantly harder challenge. The main body deploys a small security force to establish the initial contact, followed closely by a larger security force. The intent is to make the initial contact with the smallest possible force that can develop the situation. The unit making the initial contact attempts to determine the size, composition and intentions of the enemy force. The unit commander must make the initial determination whether to bypass the enemy, avoid contact (if possible), engage directly, or assist the effort of the main body.

CADET uses rules to determine the actions of the security elements. Each individual element follows the rules to decide its actions on contact. These actions ripple through the team. For instance, if the lead security element encounters a particularly strong enemy force that meets the criteria for an attack by the main body, the lead security element will:

- Engage the enemy in direct fire.
- Determine the best route and point for employment for the following security body.
- Determine the possible routes for the main body attack for consideration by the commander.
- Secure the flank opposite the following security body.

The ability to derive the tasks of the subordinate elements as a result of rules-based task expansion and situational analysis is a critical aspect of CADET's planning function. In a coalition environment, this capability helps provide an objective basis for systematically identifying and allocating tasks to assets of multiple members.

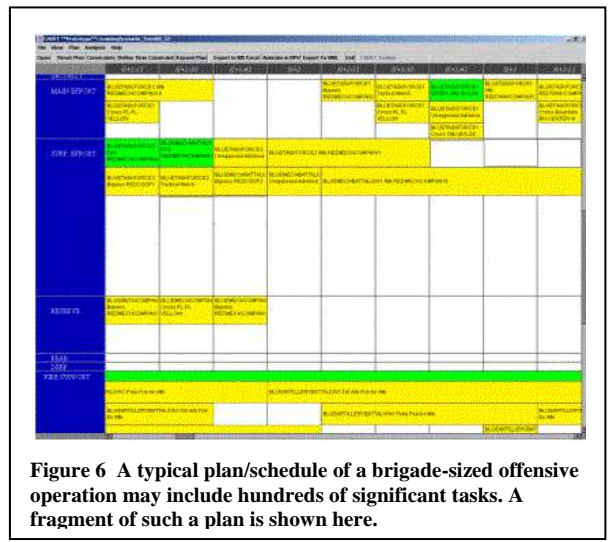

**Figure 6** A typical plan/schedule of a brigade-sized offensive operation may include hundreds of significant tasks. A fragment of such a plan is shown here.

## 3. Technical Approach

Let us consider briefly how CADET addresses some of the technical challenges implicit in the capabilities discussed above.

The integration of planning and scheduling is achieved via an algorithm for tightly interleaved incremental planning and scheduling. The HTN-like planning step produces an incremental group of tasks by applying domain-specific "expansion" rules to those activities in the current state of the plan that require hierarchical decomposition. The scheduling step performs temporal constraint propagation (both lateral and vertical within the hierarchy) and schedules the newly added activities to the available resources and time periods (Kott, 1992; Sadeh, 1996; Kott, Ground and Budd, 2002).

The same interleaving mechanism is also used to integrate incremental steps of routing, attrition and consumption estimate. For estimates of attrition, we developed a special version of the Dupuy algorithm (Kott, Ground and Langston, 1999) that was calibrated with respect to estimates of military professionals, US Army officers. This attrition calculation can be replaced with other methods, when employed in a coalition environment.

The adversarial aspects of the planning-scheduling problem are addressed via the same incremental decomposition mechanism. In particular, the tool automatically infers (using its knowledge base and using the same expansion technique used for HTN planning) possible reactions and counteractions, and provides for resources and timing necessary to incorporate them into the overall plan. In effect, this follows the military action/reaction/counter-action analysis.

In spite of significant functionality, the algorithms of CADET provide high performance. On a modern but not exceptionally fast laptop, a typical run – generation of a complete detailed plan from a high-level COA – takes about 20 seconds. With the coalition planning process taking longer than single-nation planning, which is already considered too slow, the ability to perform multiple, rapid iterations of computerized planning is very important (Riscassi, 1993).

The knowledge base of CADET is structured for simplicity and low cost. In practice, the most expensive (in terms of development and maintenance costs) part of the KB is the rules responsible for expansion (decomposition) of activities. CADET includes a module for KB maintenance that allows a non-programmer to add new units of knowledge or over-write the old ones. This is critical in a coalition environment, where the knowledge base must be rapidly extended in field conditions, to accommodate assets and rules associated with new coalition members.

From the perspective of integration with other systems, the rigorous separation – both architectural and conceptual - of problem solving components from user interaction mechanisms, allows for integration with a variety of user-interface paradigms and systems. The extensive use of XML enables simple, inexpensive integration with a variety of heterogeneous systems, a significant advantage in environments where members of a coalition bring with them a variety of systems (Thomas, 2000).

## 4. Experimental Comparisons – CADET vs. Manual Approaches

A recent experiment, one of several series (Rasch, Kott and Forbus, 2002; Kott, Ground and Budd, 2002), involved five different scenarios and nine judges (active duty officers of US military, mainly of colonel and lieutenant colonel ranks). The five scenarios were obtained from several exercises conducted by US Army. The scenarios were all brigade-sized and offensive, but still differed significantly in terrain, mix of friendly forces, nature of opposing forces, and scheme of maneuver. For each scenario/COA we were able to locate the COA sketches assigned to each planning staff, and the synchronization matrices produced by each planning staff. The participants, experienced observers of many planning exercises, estimated that these typically are performed by a team of 4-5 officers, over the period of 3-4 hours, amounting to a total of about 16 person-hours per planning product.

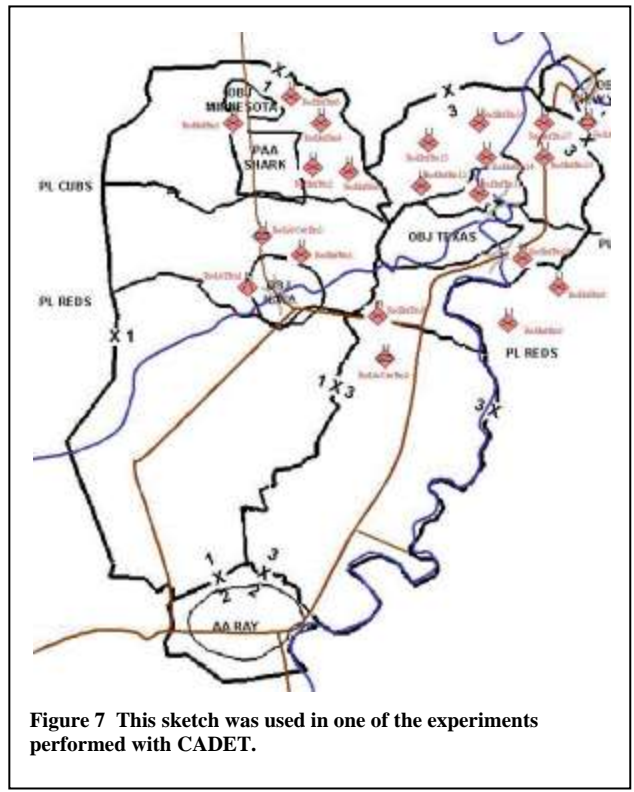

**Figure 7 This sketch was used in one of the experiments performed with CADET.**

Using the same scenarios and COAs, we used the CADET tool to generate a detailed plan and to express it in the form of a synchronization matrices. The matrices were then reviewed and edited by a surrogate user, a retired US Army officer. The editing was rather light – in all cases it involved changing or deleting no more than 2-3% of entries on the matrix. This reflected the fact that CADET is not expected to be used purely automatically, but rather in collaboration with a human decision-maker. The time to generate these products involved less than 2 minutes of CADET execution, and about 20 minutes of review and post-editing, for a total of about 0.4 person-hours per product. The resulting matrices were transferred to the Excel spreadsheet and given the same visual style at that of human-generated sets.

The products of both the CADET system and of human staff were organized into packages and submitted to the nine judges. Each package consisted of a sketch, statement, synchronization matrix and a questionnaire with grading instructions. The judges were not told whether any of the planning products were produced by the traditional manual process or with the use of any computerized aids.

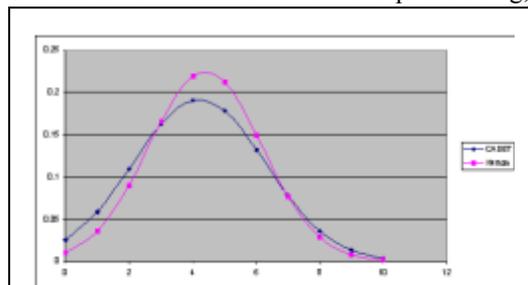

**Figure 8 The results of experiments approximated as normal distributions: the judges were asked to grade the products of CADET and manual process on a scale of 0 to 10.**

To avoid evaluation biases, assignments of packages to judges were fully randomized. Each judge was asked to evaluate four packages. Each judge was asked to review a package and grade the products contained in the package.

The results demonstrate very little difference between CADET's and human performance. In particular, based on the mean of grades, CADET lost in two of the five scenarios, won in two, and one was an exact draw. Taking the mean of grades for all five scenarios, CADET earned 4.2, and humans earned 4.4, with the standard deviation of about 2.0, a very insignificant difference.

The basic conclusion is clear: the judges gave CADET-produced products (which took typically about 20 minutes to produce) essentially the same level of grades as to the human-produced products (which took on the order of 16 person-hours to produce).

## 5. The Coalition Perspective: Conclusions and Future Work

A tool like CADET is applicable to a planning process where the planners are tasked with rapid synchronization of assets and actions of heterogeneous assets belonging to multiple organizations from multiple nations and services, potentially with distinct doctrines. The assets that enter CADET's problem solving process do not need to belong to one nation or service. Instead, each asset, e.g., a unit of force, could have its own doctrine, capabilities and rules of engagement (ROE). It should be added that coalition planning often occurs at a higher level of abstraction, dealing with problem that are not limited to a single battle but rather a long-term campaign that must consider a broad range of actions and effects – political, military, economic, social, ideological, etc. (Kott and Corpac, 2007) Such breadth and complexity of a problem go well beyond the scope of CADET.

The version of the HTN planning paradigm employed by CADET allows that a composite task is decomposed into lower-level subtasks by multiple different methods where the appropriate one is selected depending on which coalition resource would be applicable or assigned to the task. The object-oriented representation of tasks allows economical representation of nation-specific doctrinal variations applicable to the planning and execution of the task.

The integrated planning-scheduling process allows the tool to pick and choose the best coalition force, based on applicability, availability and ROE even if the assets belong to different nations. The mechanisms for flexible human intervention provide opportunities for adjusting system's choices and guiding a system in selecting proper matches between multi-force tasks and resources.

Officers belonging to different nations will need to modify or augment the knowledge base in accordance with their nation's specific doctrine. To this end, the CADET suite includes a mechanism that allows an end-user, a non-programmer, to enter definitions and rules of tasks and store them in a user-specific segment of the knowledge base. Officers can define the knowledge in the field, in real time, even while the coalition is forming and the members are defining the constraints and rules of their participation.

Coalition operations also highlight the need for a tool like CADET to allow collaborative, distributed work. Staff officers will function over geographically dispersed areas, using their adapted version of CADET on a highly portable personal computing device. Each officer on the staff uses his copy of CADET to perform a slice of the overall planning task by (a) considering the partial plans that arrive electronically from other collaborating officers; (b) making reasonable assumptions when actual partial plans are not available; (c) issuing its own partial plans to other officers and highlighting inconsistencies, if any. Although currently CADET functions as a single-user tool, we are considering plans to extend the tool for multi-user, coalition-staff operations.

At this time, CADET shows promise of reaching the state where a military decision-maker, a commander or a staff planner, uses it routinely as part of an integrated suite of tools to perform planning of tactical operations, to issue orders, and to monitor and modify the plans as the operation is executed and the situation evolves. It is not too far-fetched to suggest that such a tool may provide an 80% solution, under most situations, in a fraction of the time required for comparable manual staff planning products.

However, CADET's current state of capabilities also points toward the key gaps that must be overcome to realize the full potential of such tools in coalition warfare:

The coalition planning process is particularly demanding on effective human-machine interfaces that can be used in spite of staff members' differences in training and procedures. Such interfaces remain elusive, especially for complex, multi-dimensional information such as plans and execution of military operations, in high-tempo, high-stress, physically challenging environments. Today's common paradigms – map-based visualizations of spatial information and synchronization matrix for temporal visualization – are not necessarily the best approach, and different methods ought to be explored.

Presentation of the CADET's products requires qualitatively different user interfaces and visualization mechanisms. Our experiments suggests that users had difficulties comprehending the synchronization matrix generated by the computer tool, even though it was presented in a very conventional, familiar manner. Perhaps, the synchronization matrix functions well only as a mechanism for short-hand recording of one's own mental process and is not nearly as useful when used to present the results of someone else's, e.g., a computer tool's, reasoning process.

Ongoing work on CADET technology focuses on closing these critical gaps.

## Acknowledgements

The work described in this paper was supported by funding from US Army CECOM (DAAB 07-96-C-D603 - CADET SBIR Phase I, DAAB 07-97-C-D313 – CADET SBIR Phase II and DAAB 07-99-C-K510 - CADET Enhancements); DARPA (DAAB07-99-C-K508 Command Post of the Future); and TRADOC BCBL-H (GS-35-0559J/DABT63-00-F-1247). TRADOC BCBL-L provided additional funding under DAAB 07-99-C-K510.

Dr. L. Rebbapragada of Army CECOM contributed significantly to the concept development along with technical monitoring of the project and valuable guidance. MAJ R. Rasch of the US Army BCBL-L provided important advice. John Langston, LtCol, US Army (ret.), contributed significantly to the understanding of the functional domain and the conceptualization of the Conflict Resolution Model.

Kott, A., Corpac, P. S. (2007) "COMPOEX Technology to Assist Leaders in Planning and Executing Campaigns in Complex Operational Environments," 12th International Command and Control Research and Technology Symposium, 2007

Rasch, R., Kott, A. and Forbus, K. D. (2003) "Incorporating AI into military decision making: an experiment." IEEE Intelligent Systems 4 (2003): 18-26.

Riscassi, R. W. (1993) "Principles for Coalition Warfare", Joint Force Quarterly, Summer, 1993, pp. 58-71.

Sadeh, N.M. and Kott, A. (1996) "Models and Techniques of Dynamic Demand- Responsive Transportation Planning," Technical Report CMU-RI-TR-96-09, The Robotics Institute, Carnegie Mellon University, Pittsburgh, PA, May 1996

Thomas, T. L. (2000) "Kosovo and the Myth of Information Superiority", *Parameters: US Army War College*, Spring 2000, Vol 30, Issue 1.

Wass de Czege, H. and Biever, J. D. (2001) "Six Compelling Ideas On the Road to a Future Army," Army Magazine, Vol.51, No.2, February 2001, pp. 43-48
10